# Dynamic Teaching in Sequential Decision Making Environments

**Thomas J. Walsh**
Center for Educational Testing and Evaluation
University of Kansas
Lawrence, KS 66045

**Sergiu Goschin**
Department of Computer Science
Rutgers University
Piscataway, NJ 08854

## Abstract

We describe theoretical bounds and a practical algorithm for teaching a model by demonstration in a sequential decision making environment. Unlike previous efforts that have optimized learners that watch a teacher demonstrate a static policy, we focus on the teacher as a decision maker who can dynamically choose different policies to teach different parts of the environment. We develop several teaching frameworks based on previously defined supervised protocols, such as Teaching Dimension, extending them to handle noise and sequences of inputs encountered in an MDP. We provide theoretical bounds on the learnability of several important model classes in this setting and suggest a practical algorithm for dynamic teaching.

## 1 Introduction

In situations where one agent teaches another, the AI community has largely focused on teachers that demonstrate a static policy to the learning agent [Abbeel and Ng, 2005; Walsh et al., 2010]. However, demonstration of even an optimal policy is often not the best way to teach. For instance, consider a learner being trained to play billiards by observing a near-optimal player. Since a hallmark of good play is simplifying the next shot, an optimal player will likely only demonstrate easy shots. But if we really want to teach the learner to shoot pool, we need to show it difficult shots and novel situations that the optimal policy will rarely encounter. More generally, teachers can improve learning efficiency by showing highly informative examples that a static policy might not often encounter. In this paper, we show how to cast the problem of optimal teaching as a decision problem in its own right, and provide bounds on the teachability of several important model classes in Reinforcement Learning (RL) [Sutton and Barto, 1998].

Our approach uses lessons from the supervised-learning community, which has established a number of frameworks for quantifying the teachability of a domain class. Specifically, we extend the classical *Teaching Dimension* framework (TD) [Goldman and Kearns, 1992] and the recently described *Subset Teaching Dimension* (STD) [Zilles et al., 2011], which quantify the teachability of deterministic hypothesis classes. We adapt these frameworks for the type of data encountered in RL, particularly handling noise and sequential inputs. This allows us to characterize the teachability of several RL concept classes such as Bernoulli distributions, $k$-armed bandits and DBNs in a supervised setting.

These supervised algorithms and analyses form the foundation for our teaching algorithms in the sequential decision making setting, where teachers are constrained in the examples they can select by the dynamics of their environment and their current state. We show how to cast the problem of optimal teaching as a decision problem in its own right and that exact optimization is often intractable. We then use our results from the supervised setting to construct approximations to the optimal teaching strategy that are successful in practice. The end result is a general algorithm for a teacher in a Markov Decision Process (MDP) [Puterman, 1994] that, in contrast to work on demonstrating static policies, dynamically chooses highly informative examples to teach the model.

## 2 Alternate Approaches

We now describe competing protocols for teaching in the RL setting. We will cover previous work on teaching in the supervised setting in later sections.

Static policies for teaching RL agents have been studied in Inverse Reinforcement Learning (IRL) [Abbeel and Ng, 2004, 2005] and Apprenticeship Learning [Walsh et al., 2010]. However, in both of these settings, the emphasis is on better learning algorithms to interpret a static teaching policy repeated over many episodes. By contrast, we are providing algorithms for the teaching side so that the teacher can show trajectories built to teach, not just to demonstrate. Also, learning efficiency with static demonstrators is measured by either the time needed to achieve the same policy as the teacher (IRL) or perform as well or better than the teacher (Apprenticeship). In our current work we will focus on algorithms to teach the *entire model* of a given environment rather than a specific policy. We note this is a different problem because the learner cannot rely on a policy bias towards the teacher demonstrations.

There has been significant work on human teachers guiding RL agents, including by direct demonstration [Argall et al., 2009] and providing shaping rewards [Knox and Stone, 2010]. Most of these works focus on either optimizing the learner for human inputs or studying how different human interfaces or reward signals impact the learner. An area of future work is studying how similar our optimal teaching policies are to human teachers, as results indicate these behaviors may be very different [Khan et al., 2011].

In some domains, teaching a model may be less efficient than directly teaching the optimal policy or value function. For instance, in the bandit setting, teaching the model (the payout of all arms) requires pulling each arm a number of times, but teaching the optimal policy directly might require (if the learner knows only one arm is optimal) only pulling the optimal arm once. This situation is similar to behavioral cloning [Bain and Sammut, 1995; Khardon, 1999], where supervised learning techniques are employed in the policy space, though the focus of that field is on learning algorithms, not teaching. We focus on teaching a model because in large structured domains, the number of parameters in the model is usually much smaller than the number of parameters needed to encode the value function or policy, though our techniques could be extended to teach in the policy space.

## 3 Teachers and Learners

In this section, we begin by describing several teaching frameworks from the supervised-learning literature. We then describe modifications to these protocols to accommodate teaching in an RL setting. Specifically, we need to account for (1) noise in the labels and (2) sequences of instances rather than sets. We deal with the first problem in this section and then describe several domain classes that can be taught in this noisy supervised model.

### 3.1 Supervised Teaching Frameworks

We now describe two frameworks that can be used to measure the sample complexity of teaching in a supervised setting where a teacher is picking the examples. These frameworks differ considerably in how much the teacher and learner can infer about one another. First, we consider the classical *Teaching Dimension* model [Goldman and Kearns, 1992] where the teacher is providing helpful examples to the learner, but the learner is unaware that it actually has a teacher.

**Definition 1.** *The* Teaching Dimension *of a concept class* $\{c \in C\} : X \mapsto Y$ *over instance space $X$ and labels $Y$ is $TD(C) = \max_{c \in C} \min_{|S|} (Cons(S,C) = \{c\})$, with $Cons$ being the concepts in $C$ consistent with $S$. This makes $S$ a* teaching set $TS(c,C)$ *for $c$.*

Intuitively, TD represents the minimum number of examples needed to uniquely identify any $c \in C$. Because the learner may not know it is being taught and the teacher does not know what learner it is providing samples to, TD teachers may not act optimally for a specific learner. However, a natural extension is a protocol where the learner and teacher both understand they are interacting with a shared goal (for the learner to realize a concept). Such notions were formalised first by Balbach and Zeugmann [2009] who devised a protocol where the learner would make inferences about hypotheses that the teacher was clearly not trying to teach based on the size of the teaching set. This reasoning between the teacher and learner was further formalised in the *Subset Teaching Dimension* (STD) [Zilles et al., 2011] where the learner and teacher both assume the other is optimal.

**Definition 2.** *The* Subset Teaching Dimension *(STD) of a hypothesis class $\{c \in C\} : X \mapsto Y$ ($STD(C)$) is equal to the worst case number of samples needed to teach a learner when both the teacher and the learner know that the other is performing their task optimally. This is done by iteratively constructing Subset Teaching Sets $STS^i(c,C)$ starting from $STS^0(c,C) = TS(c,C)$, such that $STS^i \subseteq STS^{i-1}$ and $STS^i(c,C)$ is not a subset of $STS^{i-1}(c',C)$ for $c' \neq c$ (a "consistent subset").*

Intuitively, STD captures the fixed point behavior of a learner and teacher that recursively assume the other is doing their job optimally. One can think through the process of reaching this fixed point iteratively starting with the teaching sets for all concepts in the original TD framework, denoted $STS^0 = TS$. In the first step of reasoning, the learner assumes it is being taught using $STS^0$, and therefore it can assume the instances it will see will come from an optimal teaching set. This in turn allows it to notice when it has seen subsets of instances that must be leading to one (and only one) set from STS$^0$. The next step of reasoning goes back to the teacher, who infers it can teach using one of these subsets from STS$^1$. The process repeats until there are no more changes to STS. The uniqueness and reachability of this fixed point is detailed by Zilles et al. While the reasoning process may be complicated, the resulting behavior can be quite intuitive. For instance, in the original TD setting, when teaching any singleton concept over $n$ variables with the empty set also a possibility, $n$ negative instances are needed to show the empty hypothesis is correct. However in STD, all hypotheses can be taught using just one example (positive for the singleton, negative for empty).

### 3.2 Teaching Frameworks for Noisy Concepts

In this section, we extend the protocols above to the setting where the concept being learned is noisy. We begin by defining a stochastic concept and unordered collection (a "set" that may contain duplicates):

**Definition 3.** *A* stochastic concept *$c : X \mapsto D(Y)$ maps each possible instance $x \in X$ to a distribution over the label space ($D(Y)$). An* unordered example collection *$U$ of a concept $c$ consists of (potentially overlapping) pairs of inputs and labels $\{x_0, y_0...x_n, y_n\}$ with each $y_i$ drawn from $c(x_i)$.*

Next, we need to address the consistency of the learner, which can no longer be expected to predict the exact label of every instance. Instead, we consider learners that predict a distribution of labels for a given input.

**Definition 4.** *A* distribution consistent learner *with parameters $\epsilon$ and $\delta$ makes predictions for input $x$ based on the current unordered collection $U$ in the form of a predicted distribution $D'(x)$ over the label space $Y$. Consistency means $||\hat{D}(x) - D'(x)||_{TV} \leq \epsilon$ with probability $1 - \delta$, where $|| \cdot ||_{TV}$ is the total variation (half the L1-norm) over the labels and $\hat{D}$ is the distribution observed in $U$ (in most cases $\hat{D}$ is the distribution determined by the maximum-likelihood estimate).*

Now we can extend the notion of a teaching set from Definition 1 in three ways. First, instead of sets, we consider collections with duplicates. Second, we assume the teacher does not control the label associated with a given input, but can see the label (produced by the true concept) once an instance is added to the teaching collection. Therefore, the teacher can choose instances one at a time to add to the collection and always knows how all the current examples have been labeled. In practical terms, the teacher will then be able to choose when to execute a *stop* action to declare the collection finished *but* notice that the learner will not be aware of the order in which examples were added.

Finally, we say such a collection is a teaching collection (denoted $TS$ in an abuse of notation) if any distribution consistent learner must have $||D(x) - \hat{D}(x)|| \leq \epsilon$, with probability $1 - \delta$ after seeing that sequence.

We now have all of the components to define the noisy teaching dimension (NTD):

**Definition 5.** *The* noisy teaching dimension (NTD) *with parameters $\epsilon$ and $\delta$ of a stochastic concept class $C$ (NTD$(C)$) is the maximum size minimum teaching collection $\tau \in TS(c)$ over all concepts $c \in C$. That is, $\max_c \min_{\tau \in TS(c)} |\tau|$*

Next, we consider the case of a Noisy Subset Teaching Dimension (NSTD), which will extend STD. Here, we need to redefine the notion of subset used in the original STD to account for duplicates and the incremental construction of the teaching collection. To do so we introduce the *consistent subcollection* relation:

**Definition 6.** *A collection $U'$ is a* consistent subcollection *of another collection $U$ ($U' \subseteq U$) if (1) every element of $U'$ is mapped by a function to an element of $U$ and (2) the range and probability of distributions represented by $U$ and $U'$ are the same.*

Replacing the original "consistent subset" requirement from Definition 2 with a requirement that each iteration of reasoning must produce a consistent subcollection and assuming the learner is distribution consistent gives us a full definition of NSTD. In most cases the original NTD collection (NSTD$^0$) will not be improved upon in the worst case, but in many cases the teachers ability to *stop* constructing the collection will cause a significant decrease in the *expected* teaching time for NSTD over NTD (see Theorems 2 and 3). While the definitions above assumes the teacher's construction ordering is hidden from the learner, we will later see the sequential setting reveals this order and allows stronger inference.

## 4 Concept Classes

We now describe several concept classes that are relevant for RL agents. Our analysis in this section is carried out in the supervised setting.

### 4.1 Monotone Conjunctions

Conjunctions of terms are a simple but important model for pre-conditions of actions and simple conditional outcomes. A monotone conjunction over an input space of $n$ boolean terms is labeled 1 if all of the relevant variables are 1, and 0 otherwise. In the TD setting, the complexity of learning a monotone conjunction is $O(n)$ [Goldman and Kearns, 1992], but the types of examples are slightly different than in the classical mistake-bound [Littlestone, 1988] case, which only requires positive examples. This is because in TD we cannot assume anything about how the learner defaults in its predictions. Instead, the TD teacher should present the most specific positive example (only relevant variables set to 1), and then negative examples with each *relevant* variable alone set to 0.

In the STD protocol, the bound actually becomes 1 [Zilles et al., 2011]. The optimal STD strategy is to show the positive example from the TD case because the learner can infer all of the other negative examples. These results provide intuition about how preconditions or conditional effects can be taught in the sequential setting, a topic we return to in Section 5.3.

### 4.2 Coin Learning

One of the fundamental noisy hypothesis classes in model-based RL is the Bernoulli distribution learned through observation outcomes, (i.e. determining the bias ($p^*$) of a coin by observing flips). In the NTD case, because the teacher has to provide samples that may be interpreted by any type of consistent learner, the following strategy produces the optimal teaching set with high probability.

**Theorem 1.** *In the NTD setting, the proper teaching strategy is to collect $m = H(\epsilon, \delta) = \frac{1}{2\epsilon^2} ln(\frac{2}{\delta})$ samples of the Bernoulli distribution and then say* stop.

*Proof.* This strategy produces enough samples for learners that requires $m$ samples before they make any predictions (e.g. KWIK learners [Li et al., 2011]). Suppose a learner existed that made inaccurate ($> \epsilon$ error) predictions with probability $> \delta$ after seeing these samples with empirical mean $\hat{p}$. Then by Definition 4 this learner would be inconsistent because Hoeffding's inequality states $Pr[|\hat{p} - p^*| > \epsilon] < \delta$. □

This is not surprising since this is also the sample complexity as an autonomous coin learner [Li et al., 2011]. However, in NSTD, the teacher has a significant impact. The following two theorems lay out (1) the reasoning that leads to the different behavior and (2) the sample complexity of coin learning in this protocol.

**Theorem 2.** *The general NSTD teaching policy for teaching the probability of a weighted coin is to flip the coin at most $H(\epsilon, \delta)$ times, but the teacher can* stop *building the collection whenever the empirical mean $\hat{p}$ of the coin's bias is within $\epsilon/2$ of the true probability.*

*Proof.* After processing a teaching collection of some size $m$, a learner can construct a confidence interval (using Hoeffding's inequality), denoting the upper and lower limits of $p^*$ (the true probability of heads) with probability $1 - \delta$. We call this region the *consistent* region and the hypotheses outside of this region are deemed *inconsistent*. In addition, we can define the empirical mean $\hat{p} = \sum_{i=1}^{m} y_i/m$ and a region $[\hat{p} - \frac{\epsilon}{2}, \hat{p} + \frac{\epsilon}{2}]$ that we call the *instantaneous* region.

In NTD (NSTD$^0$) we just saw the teacher's only recourse is to provide $m = H(\epsilon, \delta)$ samples. For NSTD$^1$, where the learner is aware it is being taught, suppose the teacher presents a collection of size $m' < m$ to the learner. If $p^*$ is in the inconsistent region, then the label set has probability $< \delta$, so the teacher was correct in stopping early. If $p^*$ was in the consistent region but *not* in the instantaneous region, then the teacher would not have stopped construction, because treating this as a consistent subcollection will not guarantee the learner picks a distribution within $\epsilon$ of $p^*$. Therefore, in the case where the teacher stops the construction while the consistent region is still larger than the instantaneous region, the teacher's current collection is a consistent subcollection of a possible NTD collection and the learner can pick any hypothesis from the instantaneous region and be correct. □

We will now prove that the expected time for teaching a distribution in the NSTD framework is significantly smaller than the standard Hoeffding bound ($H(\epsilon, \delta)$) that determines the number of samples needed to learn in the NTD protocol. We believe the result is interesting in its own right as it describes a useful fact about the expected time the empirical average needs to first hit an interval of interest centered at the true expected value of a random variable.

The key technique we will use is formulating the evolution of the empirical average as a random walk and reducing the problem of hitting the desired interval to the problem of computing the mean first passage time through a fixed barrier. The models we introduced so far use Bernoulli as the standard example of learning a noisy concept. In the interest of technical brevity, we will actually prove the result for the case where the noise model is a Normal distribution. The reason is that the proof is more insightful for a continuous distribution and it is known in the literature [Redner, 2001; Feller, 1968] that the first passage time properties in the continuous case approximate the discrete version well.

**Theorem 3.** *Given a Normal distribution with unknown mean $p$, and the ability to sample from it, the expected number of samples a teacher needs to teach $p$ in the NSTD protocol scales with $O(\frac{1}{\epsilon})$ (the variance is considered known).*

*Proof.* The above theorem statement is equivalent to stating that the expected first time the empirical av-erage of a sequence of i.i.d. samples from a $D = Normal(p, 1)$ distribution hits the interval $[p-\epsilon, p+\epsilon]$ is $O(\frac{1}{\epsilon})$. Let $X_i, i = 1, 2, ...$ be i.i.d. samples from distribution $D$ and let $S_t = \sum_{i=1..t} X_i$ be a random walk on $\mathbb{R}$ (with step values sampled from $D$). The empirical average at any time $t > 0$ is of course $\hat{X}_t = \frac{S_t}{t}$. Let $Y_i = X_i - p$ (thus $Y_i \sim Normal(0, 1)$) and let $W_t = \sum_{i=1..t} Y_i$. We can thus write $S_t = pt + W_t, \forall t > 0$.

Proving a bound on the expected time it takes $\hat{X}_t$ to first hit interval $[p - \epsilon, p + \epsilon]$ is equivalent to showing a bound on the expected time it takes the random walk $S_t$ to hit the dynamic interval $[pt - \epsilon t, pt + \epsilon t]$. Let $A_t = pt + \epsilon t$ be a process that encodes the time evolution of the upper bound of the dynamic interval.

We will first focus on the expected time it takes for the random walk to first be inside the interval from the perspective of the upper bound. Let $B_t = A_t - S_t = \epsilon t - W_t = \epsilon t + W_t$ where the last equality follows from $W_t$ being a symmetric stochastic process around 0. Since we are looking for the expected time $t$ that $A_t$ first becomes larger than $S_t$, this is equivalent to asking what is the expected first time that $B_t$ hits the origin if it first starts on the negative side (if it starts on the positive side, this first time will be 1).

We will now approximate the discrete-time stochastic process $B_t$ by a continuous-time process with the goal of getting a qualitative result about the expected mean time. This technique is commonly used to study first passage-time properties of discrete time random processes (see for instance the Integrate-and-Fire Neuron model in section 4.2 in [Redner, 2001]). Viewed from this perspective, $B_t = \epsilon t + W_t$ is actually the standard Brownian motion with positive drift $\epsilon$. But it is well known that the mean first passage time over a fixed positive constant $\alpha$ for a Brownian motion with positive drift $\epsilon$ is governed by the **Inverse Gaussian Distribution (IG)** with parameters $IG(\frac{\alpha}{\epsilon}, \alpha^2)$ [Chhikara and Folks, 1989]. The expected value of the IG distribution is $\frac{\alpha}{\epsilon}$ and if we take $\alpha = 1$ we get that the expected first time that $B_t$ will become larger that 1 is $\frac{1}{\epsilon}$. Since the expected time to first hit the origin if started on the negative side is naturally upper bounded by the expected time to hit 1, we get a bound on the expected time $A_t$ needs to first 'catch up' with the random walk $S_t$.

Symmetrically we can show that the expected time for $S_t$ to become larger than the lower bound of the dynamic interval (the process $pt - \epsilon r$) is also $\frac{1}{\epsilon}$. $\square$

Figure 1a shows empirical validation of this result in the coin-flipping case for increasingly smaller values of $\epsilon$ (1000 runs, $\delta = .05$). While NTD grows quadratically in accordance with $H(\epsilon, \delta)$, the NSTD expected time grows only linearly in $\frac{1}{\epsilon}$.

### 4.3 $k$-armed Bandits

A natural extension of coin-learning is teaching a complete model in the $k$-armed bandit case [Fong, 1995], that is teaching the expected payout of all $k$ arms, each of which may have a different noisy (but bounded $[0, 1]$) payout function. We note again that we are teaching the full model here, not the optimal policy. Each arm can be treated as a Bernoulli distribution that needs to be learned within $\epsilon$ with probability $\delta/k$ to ensure total failure probability at most $\delta$. Note $\epsilon$ does not change with $k$ because each arm corresponds to a different input parameter $x$ (a different action). Hence for $k$ arms, the NTD solution is to pull each arm $H(\epsilon, \delta/k)$ times, giving us an NTD bound of $\frac{k}{2\epsilon^2}ln(\frac{2k}{\delta})$. For NSTD, the teacher can again pull the arms in some ordering, but can stop pulling an arm when it has either (1) been pulled $m = H(\epsilon, \delta/k)$ times or (2) its empirical average is within $\epsilon/2$ of its true payout. The expected savings over NTD is a factor of $\frac{1}{\epsilon}$ *for each arm*, so the speedup effect is actually multiplied across the arms. Figure 1b illustrates this in the bandit setting for increasing $k$ (1000 runs, $\epsilon = 1/45$, $\delta = .05$). The algorithms just described are labeled NTD-IND and NSTD-IND (for "individual" pulls). The growth of NTD-IND is actually $kln(k)$ but quickly diverges from the other approaches here, showing the increasingly better (than NTD-IND) expected performance for NSTD-IND.

While not applicable in RL, the complexity of teaching by pulling all of the arms in parallel will be informative in the next section, so we investigate it here. The goal is still to learn each arm's expected payout with $\epsilon$-accuracy, but the teacher has access to an action that pulls all of the arms at once and reports their individual payouts. NTD in this case (NTD-PAR in Figure 1b) performs $H(\epsilon, \delta/k)$ parallel pulls, saving a factor of $k$ over the individual NTD above. But in NSTD, the parallel pulls introduce a tension between the speedup from parallelizing and the previously noted speedup from being able to stop pulling an arm when its empirical mean is close to the true mean. Now if some arms are close to their empirical mean but others are not, a parallel pull may disturb the empirical means of the "learned" arms. Figure 1b shows the NSTD-PAR strategy that is forced to either perform a parallel pull or *stop*, which it only does when *all* the empirical payouts are within $\epsilon/2$ of their true payout or have been pulled $H(\epsilon, \delta/k)$ times. For small $k$, the sequential pulls are actually more efficient, but for a larger number of arms, the parallel pulls have a significant benefit despite the danger of "unlearning" an arm.

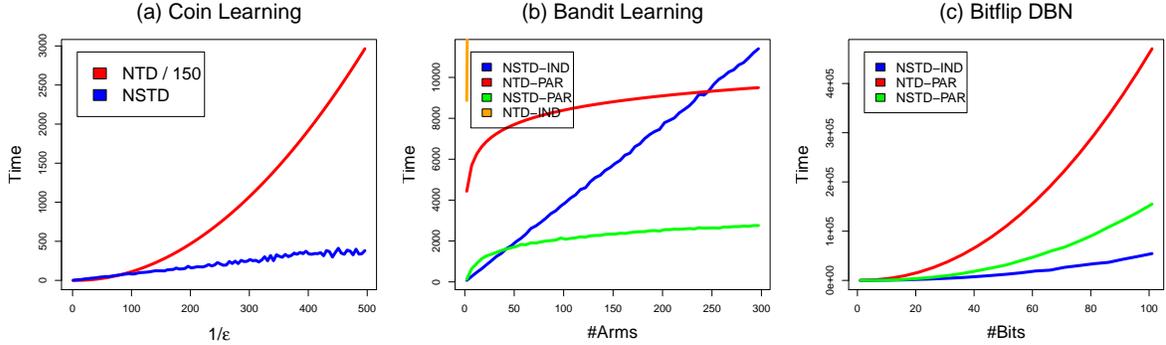

Figure 1: TD and STD comparisons for (a) coin flipping, (b) $k$-armed bandits and (c) the Bitflip DBN

### 4.4 Dynamic Bayesian Networks

We now consider Dynamic Bayesian Networks [Dean and Kanazawa, 1989], or DBNs, where multiple noisy factors are learned in parallel. This case is similar to the parallel bandit case but here because the total error ($\epsilon$) is an aggregate of the subproblem errors, the NSTD strategy will be dramatically different. A Dynamic Bayesian Network is composed of $n$ discrete valued factors. At every timestep, the next value for factor $x_i$ is determined by a probability distribution $P(x_i|\rho(x_i))$ where $\rho$ maps the factor to its $k$ "parent factors". We assume here that $k = O(1)$ and we only consider the binary case (all factors are either 0 or 1) as the bounds generalize to ordinal values from 1 to $D$ with extra terms including $D^{k+1}$ (see [Li et al., 2011]).

Mapping this back to our learning protocols, the input space is all possible configurations of the DBN factors and predictions should be made on the distribution of the factors at the next timestep. As a simple example, consider the case where every variable is the parent of only one other variable. Assume that the structure is known and the DBN is *deterministic*, so all that has to be learned is the relationship between each parent value and its child value. In the traditional mistake bound setting, where the teacher has no control over the inputs, the worst case learning bound is $O(n)$, because each example after the first one could show the same parent (input) pattern as the first, except with a single bit flipped. Because of the connection between mistake bound learnability and teaching by demonstration [Walsh et al., 2010] this is the worst-case bound for teaching this type of DBN by demonstration of a static policy. However, with our dynamic teaching protocols we can do much better. Specifically, in the TD case the sample complexity is $O(1)$, because the teacher can pick the parent values. In the first example, the teacher can pick an arbitrary setting of the factors and show the result. Then all it needs to do is show the complement of this bit string in the second example. Thus, deterministic binary DBNs with $k = 1$ are teachable with 2 examples. The STD bound is the same in this case and the generalization to multiple parents adds only a constant term.

In the stochastic setting, the consequences of teaching the factor probabilities in parallel are more complicated than the deterministic case. Unlike the bandit case, the total error in predicting the next state probability is based on the aggregate error (total variation by Definition 4) of all the subproblems. So each factor needs to be predicted with $\epsilon/n$ accuracy. This means that for an NTD teacher, we need $H(\epsilon/n, \delta/n^k) = O(\frac{n^2}{\epsilon^2} k \ln(\frac{n}{\delta}))$ samples to learn the probabilities.

The stochastic setting is even more complicated in the NSTD case, where once again there is competition between the desire to teach factor probabilities in parallel and the desire to stop showing certain conditions once they have been taught. The parallel and individual teaching styles in the bandit case manifest themselves here in the following ways. NSTD-PAR tries to teach multiple factors at once, using the same "complement" input selection as NTD but can stop providing samples whenever *all* of the factors are $\epsilon/2n$ accurate. The worst case time for this approach is bounded by NTD's bound above but again on expectation this will usually perform better than NTD. However, if one wishes to minimize expected teaching time when there is background knowledge that says certain conditions are already known, a variant of the NSTD-IND strategy is likely better. In NSTD-IND, the teacher presents inputs with only one unknown condition for a single factor, and all other factors have their parent values set to configurations matching the background knowledge. As in the bandit case, this avoids "unteaching" but sacrifices parallelization. Once the factor outcome being taught has its empirical distribution within $\epsilon/2n$ of the true probability, the algorithm moves on to the next condition. The worst-case time for NSTD-IND is $O(\frac{n^3}{\epsilon^2} k \ln(\frac{n}{\delta}))$, worse than TD and NSTD-PAR, but its expected time is only $O(\frac{n^2}{\epsilon}kln(\frac{n}{\delta}))$ since it targets individual conditions.

An empirical demonstration of these teaching strategies is shown in Figure 1c for a DBN representation of a noisy Bitflip domain [Givan et al., 2003]. The input for this DBN is simply a bit string of length $n$ and there are two actions, one to flip bit 0 and one to *shift* each bit up (with a 0 always incoming to bit 0). However, the effect of the shift is noisy for each bit: with probability $p_i$ the shift to bit $i$ is successful, and otherwise the bit retains the same value it currently has. Figure 1c shows the number of steps needed to teach all of the $p_i$'s for $\epsilon = .3$ (aggregate error) and $\delta = .05$ for the three teaching protocols described above (500 runs). NTD teaches by setting the inputs (current state) to an alternating string of 0's and 1's ending in a one, so that each bit's shift probability is observed on every step (ensuring parallel teaching). NSTD-PAR uses the same alternating bit string for every example, but stops building the collection if all of the $p_i$ values are within $\epsilon/2n$ of their true value. NSTD-IND sets up the state to have all 1's from bit 0 up to the highest bit it has not yet taught, and then all $0's$ above that. This teaches the probability of shifting each bit one at a time with deterministic outcomes for the other bits. We see here that unlike the bandit case, NSTD-IND dominates the other strategies, even as the number of bits increases. This is because the continually shrinking accuracy requirements increase the chances of the parallel strategy unteaching a factor. However, NSTD-PAR still outperforms NTD, so in situations where no background knowledge is available (for NSTD-IND) and the teacher and learner are aware of each other, this may be the preferred strategy.

Finally, we note Bitflip also showcases the benefit of teaching versus demonstration from an optimal policy. Consider Bitflip with a reward only when all bits are 1. The optimal policy is to flip if bit 0 is a 0, otherwise shift. However, this strategy produces very few useful samples because once a bit is turned to 1 in an episode, it will never be a 0 (even if the shift fails the bit retains its current value). Even if $p_i = .5$, the expected number of useful samples in an episode for bit $i$ is only 2, and the probability of more samples drops off exponentially. So the optimal *performance* policy will almost always take far more steps to teach the domain than any of the teaching protocols described above.

## 5 Teaching in an MDP

Above, we established the teachability of several RL concept classes in a supervised setting. However, we are interested in teachers acting out lessons in an MDP, where a transition function $T : X, A \mapsto Pr[X]$ governs agent movement, and therefore the teacher may not have access to all possible states at every timestep. We now extend the previously defined frameworks to the sequential setting, forcing them to handle teaching *sequences* drawn from an MDP. We then describe the teaching process in an MDP as a planning problem in its own right and present and demonstrate a heuristic solution based on our supervised learning results.

### 5.1 Sequential Teaching Frameworks

In sequential domains, the teacher may not be able to access all possible states at every timestep, but instead may have to take multiple steps (each of which will be seen by the learner) to reach its target teaching state. Hence, we now adapt the TD and STD definitions to consider *sequences* of states and actions rather than randomly accessed inputs and labels. More formally, we need each protocol to handle the following constrained teaching sequences.

**Definition 7.** *An MDP teaching sequence $MTS(c, C, M)$ for MDP $M = \langle X, A, T, R, \gamma \rangle$ and $c : X, A \mapsto Pr[X]$ or $c : X, A \mapsto \Re$ is a sequence $\langle \langle x_0 a_0 r_0 \rangle ... \langle x_T, a_T, r_T \rangle \rangle$ with each $\langle x_i, a_i, r_i, x_{i+1} \rangle$ reachable by taking action $a_i$ from $x_i$ in $M$, starting from $x_0 = s_0$. After witnessing this sequence, the $\hat{c}$ learned by a consistent learner must either be $c$ (deterministic) or distributionally consistent (Definition 4) to $c$ with high probability (stochastic concepts).*

In the definition above, $c$ is a concept that may be part of the transition or reward function (or both) of $M$. For TD, instead of considering a set of instances and labels, one now needs to consider a sequence of states and actions drawn from $M$ based on the teacher's potentially non-stationary policy $\pi$. We denote the distribution of such sequences as $MTS^\pi(c, C, M)$. Formally, we define Sequential Teaching Dimension ($TD_S$) in an MDP as:

**Definition 8.** *The sequential teaching dimension $TD_S(C, M)$ of concept class $C$ being taught in MDP $M$ with accuracy parameters $\epsilon$ and $\delta$ is the maximum of the minimum expected length of an MDP teaching sequence for any individual concept in the class, $TD_S(C, M) = \max_c \min_\pi E[MTS^\pi(c, C, M)]$, where $\pi$ is an arbitrary, potentially non-stationary, policy (any possible teaching algorithm).*

This definition is very general and using it constructively in the full stochastic case is a topic of future work, but there are cases where the optimal dynamic teaching policy is clear. For instance, when the transition function is deterministic (including cases where we are teaching a noisy reward function), the teacher simply needs to find the shortest-length teaching sequence starting from $s_0$ that contains enough samples to teach $c$ to any consistent learner. Also, heuristic solutions for approximating optimal teaching defined in this manner (e.g. Algorithm 1) can successfully teach in both the deterministic and stochastic setting.

The conversion of Zille's STD protocol to the sequential setting is even more complicated because in STD the learner makes inferences about why the teacher is showing certain instances, but now not every instance is accessible at every step. The main intuition we can use to resolve these difficulties in the case where $T$ is deterministic is to instead apply the recursive reasoning to the teaching sequences defined above. Intuitively, this should allow the learner to reason that a teacher is trying to teach a concept $c$ when the teacher chooses a path that leads to a state consistent with $c$ instead of other paths that would teach different concepts. This saves the teacher from walking the entire path. We can formalize this using the notion of an example subsequence and prefix.

**Definition 9.** *An* example subsequence $\Sigma'$ *of example sequence* $\Sigma$ *is a contiguous series of inputs and labels* $\langle x_i, y_i...x_j...y_j \rangle$ *for* $i \geq 0$, $j \leq T$. *A* prefix *($\Sigma' <_{pre} \Sigma$) is a subsequence with $i = 0$ and a* suffix *is a subsequence with $j = T$.*

Using this definition, we can replace previous definitions of subsets and collections in STD and NSTD and define a SubSequence Teaching Dimension (SSTD) in a natural way for MDPs with a deterministic transition function (see the appendix).

As a concrete example of the more powerful reasoning about sequences in SSTD and NSSTD (its noisy version), consider coin learning, but instead of predicting a probability, assume the learner only needs to predict whether the coin is biased more towards heads or tails (assuming $p^* \neq 0.5$). Without loss of generality assume the coin is biased towards tails but comes up heads on the first trial. In NSTD, because the ordering of examples is hidden to the learner (Def. 3), we need to keep flipping the coin until we have observed more tails than heads. But in NSSTD we only need one more flip and then the teacher can end teaching, no matter what the outcome of the second flip (and even though the empirical distribution will not indicate a tails bias). This is because if the coin was biased heads the teacher would not have made the second flip; it would have stopped after the first heads observation. Similar reasoning can be done when predicting $p^*$ as every choice to flip indicates $p^*$ is *not* in the previous step's instantaneous region (see Theorem 2).

### 5.2 Practical Teaching in MDPs

If teaching is to be done in an MDP, the best policy will directly optimize the $TD_S$ or $SSTD$ criteria described above. However, even for $TD_S$ in the deterministic setting, the problem of determining the exact optimal teaching sequence may be intractable.

**Theorem 4.** *Determining the optimal teaching sequence for a concept in a deterministic MDP under the conditions of Definition 8 is NP-Hard.*

*Proof.* We can encode the graph of a Traveling Salesman Problem (TSP) with integer edge costs as a deterministic MDP with unit costs by adding in the same number of "dummy" states between cities as their distance in the original graph. Then consider teaching a reward function that is known to be 0 at every non-city state but has an unknown value at each of the cities. The shortest tour for teaching such a reward function provides a solution to the original TSP. □

This is in line with previous results on the intractability of determining teaching sets in the supervised setting [Servedio, 2001]. However, tractable approximations such as Algorithm 1 are certainly possible. There we perform two approximations: first we use a greedy approximation of the supervised (random access) teaching collection construction to find a set of instances that will teach the target concept and then greedily construct a tour of all instances in this set.

---

**Algorithm 1** Heuristic teaching in an MDP

**input:** Concept $c \in C$ to be taught, MDP $M$ with start state $s_0$, Learning protocol $P$
$TS(c, C) = \emptyset$
$R = \langle s, a, r, s' \rangle \in M$ reachable from $s_0$
**while** $TS(c, C)$ is not a teaching set for $c$ in $P$ **do**
  $X = \langle s, a, r, s' \rangle \in R$ and $\notin TS(c, C)$ that teaches the most parameters of $c$
  $TS(c, C) = \{X\} \cup TS(c, C)$
**while** $TS(c, C) \neq \emptyset$ **do**
  $X' =$ closest $X \in TS(c, C)$ to current state, reachable fastest with policy $\pi$
  Demonstrate $\pi$ until $X'$ is demonstrated
  $TS(c, C) = TS(c, C) \backslash X'$

---

In the SSTD setting, the first approximation by itself could be problematic because the inference process is based on optimality assumptions shared by the learner and teacher. If the teacher uses a suboptimal teaching sequence, then the learner can make the wrong inference. So for extending SSTD or NSSTD to the practical setting, any approximations used by the teacher to construct the teaching set or the touring policy need to be shared between the teacher and the learner, allowing them to simulate each other's behavior. The motivation behind this sharing is to engender natural teacher-student interactions by having the learner and teacher share a set of common assumptions about one another (such as optimality or consistency in the original STD and TD).

Finally, we consider noise in the transition function for states and actions that are not part of the target concept being taught. In that case, we would ideally like to create a tour of instances with the shortest stochastic path length (in expectation), but this is certainly as hard as the deterministic touring case. However, we can modify the heuristic algorithm used above to simply find the state with the shortest stochastic path and then focus on reaching that state. This allows the heuristic to teach concepts in arbitrary MDPs.

### 5.3 An Example in the Taxi Domain

To demonstrate our heuristic teaching algorithm we conducted experiments with the Object-Oriented MDP encoding of a Taxi Domain [Diuk et al., 2008]. This environment consists of a square grid (5 by 5 for us) and a controllable taxi with actions {up, down, left, right}. There are also several *landmarks* in the grid, one of which initially contains a *passenger* and another of which is the passenger's *destination*. The taxi's goal is to go to the passenger, execute its *pickup* action, transport her to the destination, and then execute *dropoff*. In the OOMDP encoding of this domain, every state has a set of predicates (e.g. *WallToRight*(taxi)) associated with it, and actions have parameters stipulating which objects are in its scope(e.g. *pickup*(T,P,L) for taxi, passenger and landmark). The predicates in this domain include indicators of walls and clearness in every direction from an object, as well as predicates indicating when two objects are in the same square and when the passenger is in the taxi. Also, each action has a conjunction over the variablized predicates that serves as a *pre-condition* for this action. We focus on teaching these pre-conditions.

We perform our experiments in a deterministic Taxi domain, starting with a modified TD-style conjunction teacher as the set constructor for Algorithm 1. The modification to the conjunction learner from Section 4.1 is that instead of using the most specific positive example, it may have to use one that contains irrelevant predicates (like *WallNorth* for *putDown*) because of state-space restrictions. The teacher must show more positive examples to discredit these irrelevant predicates in addition to the negative examples (action failures) to indicate the relevant variables.

To demonstrate the benefits of STD, we used an approximation of the STD behavior described in Section 4.1 that shows the most specific state it can, then shows positive examples for all of the irrelevant variables and then stops demonstrating (since the learner can infer all other variables are relevant). Table 5.3 shows the number of steps needed to fully teach the pre-conditions of several sets of actions with all others being known. The agent in all cases starts from the middle of the grid and there are 2 landmarks in the bottom-left and top-right corners.

Table 1: Teaching steps for selected action sets in taxi

| Action(s) | TD | STD Approximation |
|---|---|---|
| pickup | 20 | 15 |
| pickup / putdown | 23 | 17 |
| movement | 37 | 27 |
| all | 63 | 45 |

Several interesting behaviors were observed, including the teacher using the landmarks in the corners as areas to gather many positive examples (because the irrelevant predicates can be dispelled en masse there). The teacher also made use of the lack of type constraints on most of the variables to create negative instances (where the pre-conditions failed) by swapping the order of arguments to indicate specific relations were relevant. These results compare favorably to previous result on teaching taxi conditions by static-policy demonstration [Walsh et al., 2010] but static policy approaches can be made to look arbitrarily bad in this situation because they will not teach subtle aspects of the domain (like how the corners work) unless they are actually on the way to the goal. By contrast, our approach ignores the current goal and focuses on being the best teacher, not the best performer.

We also experimented with a sequential Bitflip domain (Section 4.4) where all but two of the bits (the middle and one from the end) had deterministic shift outcomes. Our experiments again showed the approximations using NSTD significantly outperforming those using NTD. The average steps to teach the 10 bit version were 362.35 (NSTD-Par), 869.61 (NSTD-IND) and 14610.41 (NTD-PAR). The parallel versions were all more efficient despite flipping every other action.

## 6 Conclusions

We have extended two supervised learning frameworks (TD and STD) to handle noisy observations and sequential data. These frameworks provide efficient ways to teach several RL classes, including DBNs and Bernoulli distributions. We also presented a practical heuristic algorithm for leveraging TD and STD to perform efficient teaching in an MDP and demonstrated its effectiveness in the taxi domain. Unlike previous efforts at teaching domains with a static policy, these new algorithms actually target the individual parameters of a domain in a task-independent manner, leading to agents that truly teach, rather than just show.